# Boosting through Optimization of Margin Distributions

Chunhua Shen and Hanxi Li

*Abstract*—Boosting is of great interest recently in the machine learning community because of the impressive performance for classification and regression problems. The success of boosting algorithms may be interpreted in terms of the margin theory [1]. Recently, it has been shown that generalization error of classifiers can be obtained by explicitly taking the margin distribution of the training data into account. Most of the current boosting algorithms in practice usually optimize a convex loss function and do not make use of the margin distribution. In this work we design a new boosting algorithm, termed margin-distribution boosting (MDBoost), which directly maximizes the average margin and minimizes the margin variance at the same time. This way the margin distribution is optimized. A totally-corrective optimization algorithm based on column generation is proposed to implement MDBoost. Experiments on various datasets show that MDBoost outperforms AdaBoost and LPBoost in most cases.

## I. INTRODUCTION

Boosting offers a method for improving existing classification algorithms. Given a training dataset, boosting builds a *strong* classifier using only a *weak* learning algorithm [1], [2]. Typically, a weak (or base) classifier generated by the weak learning algorithm has a misclassification error that is slightly better than random guess. A strong classifier has a much better test error. In this sense, boosting algorithms can boost the weak learning algorithm to obtain a much stronger classifier. Boosting was originally proposed as an ensemble learning method, which depends on majority voting of multiple individual classifiers. Later, Breiman [3] and Friedman *et al.* [4] observed that many boosting algorithms can be viewed as gradient descent optimization in functional space. Mason *et al.* [5] developed AnyBoost for boosting arbitrary loss functions with a similar idea. Despite the large success in practice of these boosting algorithms, there are still open questions about why and how boosting works. Inspired by the large-margin theory in kernel methods, Schapire *et al.* [1] presented a margin-based bound for AdaBoost, which tries to interpret AdaBoost's success with the margin theory. Although the margin theory provides a qualitative explanation of the effectiveness of boosting, the bounds are quantitatively weak. A recent work [6] has proffered new tighter margin bounds, which may be useful for quantitative predictions. Arc-Gv [3], a variant of the AdaBoost algorithm, was designed by Breiman to empirically test AdaBoost's convergence properties. It is very similar to AdaBoost (only different in calculating the coefficient associated with each weak classifier) such that it increases margins even more aggressively than AdaBoost. Breiman's experiments on Arc-Gv show contrary results to the margin theory: Arc-Gv always has a minimum margin that is provably larger than AdaBoost but Arc-Gv performs worse in terms of test error [3]. Grove and Schuurmans [7] observed the same phenomenon. In the literature, much work has focused on maximizing the minimum margin [8]–[10]. Recently, Reyzin and Schapire [11] re-ran Breiman's experiments by controlling weak classifiers' complexity. They found that a better margin distribution is more important than the minimum margin. It is of importance to have a large minimum margin, but not at the expense of other factors. They thus conjectured that maximizing the average margin rather than the minimum margin may lead to improved boosting algorithms. We try to verify this conjecture in this work.

Recently, Garg and Roth [12] introduced margin distribution based complexity measure for learning classifiers and developed margin distribution based generalization bounds. Competitive classification results have been shown by optimizing this bound. Another relevant work is [13]. [13] applies a boosting method to optimize the margin distribution based generalization bound obtained by [14]. Experiments show that the new boosting methods achieve considerable improvements over AdaBoost. The optimization of this new boosting method is based on the AnyBoost framework [5]. Aligned with these attempts, we propose a new boosting algorithm through optimization of margin distribution (termed MDBoost). Instead of minimizing a margin distribution based generalization bound, we directly optimize the margin distribution: maximizing the average margin and at the same time minimizing the variance of the margin distribution.

The theoretical justification of the proposed MDBoost is that, approximately, AdaBoost actually maximizes the average margin and minimizes the margin variance.

The main contributions of our work are as follows.

1) We propose a new totally-corrective boosting algorithm, MDBoost, by optimizing the margin distribution directly. The optimization procedure of MDBoost is based on the idea of column generation that has been widely used in large-scale linear programming.

Manuscript received April 16, 2009; revised 17 November, 2009; accepted January 5, 2010. First published January XX, 2010; current version published July XX, 2010. NICTA is funded by the Australian Government's Department of Communications, Information Technology, and the Arts and the Australian Research Council through *Backing Australia's Ability* initiative and the ICT Research Center of Excellence programs.

C. Shen and H. Li are with NICTA, Canberra Research Laboratory, Locked Bag 8001, Canberra, ACT 2601, Australia; and the Research School of Information Science and Engineering, Australian National University, Canberra, ACT 0200, Australia (e-mail: {chunhua.shen, hanxi.li}@nicta.com.au).

Color versions of one or more of the figures in this paper are available online at http://ieeexplore.ieee.org.

Digital Object Identifier 10.1109/TNN.2010.XXXXXXX

2) We empirically demonstrate that MDBoost outperforms AdaBoost and LPBoost on most UCI datasets used in our experiments. The success of MDBoost verifies the conjecture in [11]. Our results also show that MDBoost has achieved similar (or better) classification performance compared with AdaBoost-CG [15]. AdaBoost-CG is also totally-corrective in the sense that all the linear coefficients of the weak classifiers are updated during the training. An advantage of MDBoost is that, at each iteration, MDBoost solves a quadratic program while AdaBoost-CG needs to solve a general convex program.[1]

Throughout the paper, a matrix is denoted by an upper-case letter ($X$); a column vector is denoted by a bold low-case letter ($\boldsymbol{x}$). The $i$th row of $X$ is denoted by $X_{i:}$ and the $i$th column $X_{:i}$. We use $\mathbf{I}$ to denote the identity matrix. $\mathbf{1}$ and $\mathbf{0}$ are column vectors of 1's and 0's, respectively. Their sizes will be clear from the context. We use $\succeq, \preceq$ to denote component-wise inequalities.

The rest of the paper is structured as follows. In Section II we present the main idea. In Section III the dual of the MDBoost's optimization problem is derived, which enables us to design an LPBoost-like column generation based boosting algorithm. We provide an experimental comparison of the algorithms on UCI data in Section IV, and conclude the paper in Section V.

## II. ALGORITHMS

Before we present our main results, we introduce some preliminary concepts. Let $\{(\boldsymbol{x}_i, y_i)\}_{i=1,\cdots,M}$ be the set of training data, where $\boldsymbol{x}_i \in \mathcal{X}$ and $y_i \in \{-1, +1\}$, $\forall i$. Let $h(\cdot) \in \mathcal{H}$ be a base/weak classifier that projects an input vector $\boldsymbol{x}$ into $[-1, +1]$. We assume that the set $\mathcal{H}$ is finite and we have $N$ possible weak classifiers. Let the matrix $H \in \mathbb{R}^{M \times N}$ where the $(i, j)$ entry of $H$ is $H_{ij} = h_j(\boldsymbol{x}_i)$. $H_{ij}$ is the label predicted by weak classifier $h_j(\cdot)$ on the training datum $\boldsymbol{x}_i$. Therefore each column $H_{:j}$ of the matrix $H$ consists of the output of weak classifier $h_j(\cdot)$ on all the training data; while each row $H_{i:}$ contains the outputs of all weak classifiers on the training datum $\boldsymbol{x}_i$.

Boosting is a typical example of ensemble learning, where multiple learners are trained to solve a single classification problem. A boosting algorithm creates a strong learner by incrementally adding weak learners to the final strong learner [2]. The weak learner has an important impact on the strong learner. In general, a boosting algorithm builds on a user-specified base learning procedure and runs it repeatedly on modified data that are outputs from the previous iterations. The final output strong classifier takes the form $F(\boldsymbol{x}) = \sum_{j=1}^{N} w_j h_j(\boldsymbol{x})$ with $w_j > 0, j = 1 \cdots N$.

The following theorem serves as the basis of the proposed MDBoost algorithm.

**Theorem 2.1.** *AdaBoost maximizes the unnormalized average margin and simultaneously minimizes the variance of the margin distribution under the assumption that the margin follows a Gaussian distribution.*

*Proof:* See appendix.                                                                      ∎

The key assumption that makes this theorem valid is that the weak learners generated by AdaBoost make independent errors over the training dataset. This assumption may not be true in practice, but could be a plausible approximation.

Mathematically, the above theorem can be formulated as:

$$\max_{\boldsymbol{w}} \ \bar{\rho} - \tfrac{1}{2}\sigma^2, \ \text{s.t.} \ \boldsymbol{w} \succeq \mathbf{0}, \mathbf{1}^\top \boldsymbol{w} = D, \tag{1}$$

where $\sigma^2$ is the unnormalized margin variance and $\bar{\rho}$ is the unnormalized average margin. Let $\rho_i$ denote the unnormalized margin for the $i$th example datum, *i.e.*,

$$\rho_i = y_i H_{i:} \boldsymbol{w}, \ \forall i = 1, \cdots, M. \tag{2}$$

In the above equations, $\boldsymbol{w}$ is the linear coefficients that weight the weak classifiers. $D$ is the sum of these linear coefficients, which needs to be determined by cross-validation. Note that $D$ is actually a trade-off parameter, which balances the *normalized* average margin and the *normalized* margin variance. The empirical margin variance can be computed as $\sigma^2 = \frac{1}{M-1} \sum_{i>j} (\rho_i - \rho_j)^2$. So we explicitly write the optimization in $\boldsymbol{\rho}$:

$$\min_{\boldsymbol{w}} \ \tfrac{1}{2(M-1)} \sum_{i>j} (\rho_i - \rho_j)^2 - \sum_{i=1}^{M} \rho_i, \ \text{s.t.} \ \boldsymbol{w} \succeq \mathbf{0}, \mathbf{1}^\top \boldsymbol{w} = D. \tag{3}$$

If we normalize the margin by setting $\mathbf{1}^\top \boldsymbol{w} = 1$, the above problem is also equivalent to

$$\min_{\boldsymbol{w}} \ \tfrac{D}{2(M-1)} \sum_{i>j} (\rho_i - \rho_j)^2 - \sum_{i=1}^{M} \rho_i, \ \text{s.t.} \ \boldsymbol{w} \succeq \mathbf{0}, \mathbf{1}^\top \boldsymbol{w} = 1, \tag{4}$$

where now $\boldsymbol{\rho}$ is the normalized margin. From this formulation, it is easy to see that $D$ balances the two terms in the cost function. Both problems are equivalent to (1). We define a matrix $A \in \mathbb{R}^{M \times M}$:

$$A = \begin{bmatrix} 1 & -\frac{1}{M-1} & \cdots & -\frac{1}{M-1} \\ -\frac{1}{M-1} & 1 & \cdots & -\frac{1}{M-1} \\ \vdots & \vdots & \ddots & \vdots \\ -\frac{1}{M-1} & -\frac{1}{M-1} & \cdots & 1 \end{bmatrix}.$$

Our optimization problem can be rewritten into a simplified version:

$$\min_{\boldsymbol{w}, \boldsymbol{\rho}} \ \tfrac{1}{2} \boldsymbol{\rho}^\top A \boldsymbol{\rho} - \mathbf{1}^\top \boldsymbol{\rho},$$
$$\text{s.t.} \ \boldsymbol{w} \succeq \mathbf{0}, \mathbf{1}^\top \boldsymbol{w} = D,$$
$$\rho_i = y_i H_{i:} \boldsymbol{w}, \forall i = 1, \cdots, M. \tag{5}$$

It is easy to see that $A$ is positive semidefinite[2]. So (5) is a convex quadratic problem (QP) in $\boldsymbol{\rho}$.

---

[1] More precisely, it is an constrained entropy maximization problem. To date, unlike quadratic programming that is a well-studied optimization problem, there are no specialized solvers for the constrained entropy optimization problem.

[2] $A$ is not strictly positive definite. Since the sum of $A$'s each row is zero, one of $A$'s eigenvalues is zero.

If we could access all the weak classifiers (the entire matrix $H$ is knew), we can solve the problem (5) using off-the-shelf QP solvers [16]. However, in many cases, we do not know $H$ before hand simply because the size of the weak classifier set could be prohibitively (or even infinitely) large. As in LPBoost [10], column generation can be used to attack this problem. Column generation was first proposed by [17] for solving some special structured linear programs with extremely large number of variables. A comprehensive survey on this technique is [18]. The general idea of column generation is that, instead of solving the original large-scale problem (master problem), one works on a restricted master problem with a reasonably small subset of variables at each step. The dual of the restricted master problem is solved by conventional convex programming, and the optimal dual solution is used to find the new variable to be included into the restricted master problem. LPBoost is a direct application of column generation in boosting. For the first time, LPBoost shows that in a linear program framework, unknown weak hypotheses can be learned from the dual although the space of all weak hypotheses is infinitely large. This is the highlight of LPBoost. This idea can be generalized to solve convex programs other than linear programming problems.[3] We next derive the dual of (5) such that a column generation based optimization procedure can be devised.

## III. THE DUAL OF MDBOOST

The dual of a convex program always reveals some meaningful properties of the problem. We show that MDBoost is in fact a regularized version of LPBoost. The Lagrangian of (5) is

$$\underbrace{L(\boldsymbol{w}, \boldsymbol{\rho}, \underbrace{\boldsymbol{u}, r, \boldsymbol{q}}_{\text{dual}})}_{\text{primal}} = \tfrac{1}{2}\boldsymbol{\rho}^\top A\boldsymbol{\rho} - \mathbf{1}^\top \boldsymbol{\rho} + r(\mathbf{1}^\top \boldsymbol{w} - D)$$
$$- \boldsymbol{q}^\top \boldsymbol{w} + \sum_{i=1}^M u_i(\rho_i - y_i H_{i:}\boldsymbol{w}), \quad (6)$$

with $\boldsymbol{q} \succcurlyeq \mathbf{0}$. The infimum of $L$ w.r.t. to the primal variables is

$$\inf_{\boldsymbol{\rho}, \boldsymbol{w}} L = \inf_{\boldsymbol{\rho}} \left[ \tfrac{1}{2}\boldsymbol{\rho}^\top A\boldsymbol{\rho} + (\boldsymbol{u} - \mathbf{1})^\top \boldsymbol{\rho} \right]$$
$$+ \inf_{\boldsymbol{w}} \left[ (r\mathbf{1}^\top - \boldsymbol{q}^\top - \sum_{i=1}^M u_i y_i H_{i:})\boldsymbol{w} \right] - Dr. \quad (7)$$

Clearly, $r\mathbf{1}^\top - \boldsymbol{q}^\top - \sum_{i=1}^M u_i y_i H_{i:} = \mathbf{0}$ must hold in order to have a finite infimum. Therefore, we have

$$\sum_{i=1}^M u_i y_i H_{i:} \preccurlyeq r\mathbf{1}^\top. \quad (8)$$

For the first term in $L$, its gradient must vanish at the optimum:

$$\frac{\partial \left[ \tfrac{1}{2}\boldsymbol{\rho}^\top A\boldsymbol{\rho} + (\boldsymbol{u} - \mathbf{1})^\top \boldsymbol{\rho} \right]}{\partial \rho_i} = 0, \forall i. \quad (9)$$

This leads to $\boldsymbol{\rho} = -A^{-1}(\boldsymbol{u} - \mathbf{1})$; and the infimum is $-\tfrac{1}{2}(\boldsymbol{u} - \mathbf{1})^\top A^{-1}(\boldsymbol{u} - \mathbf{1})$.

By putting the above results together, the dual is

$$\max_{r, \boldsymbol{u}} \; -Dr - \tfrac{1}{2}(\boldsymbol{u} - \mathbf{1})^\top A^{-1}(\boldsymbol{u} - \mathbf{1}), \text{ s.t. (8)}. \quad (10)$$

<hr>

[3]Nevertheless, for linear programs, the optimal solution always lies at a vertex and column generation solves the program exactly. For general large-scale convex programs, only an approximate solution can be found.

We can reformulate (10) as

$$\min_{r, \boldsymbol{u}} \; r + \tfrac{1}{2D}(\boldsymbol{u} - \mathbf{1})^\top A^{-1}(\boldsymbol{u} - \mathbf{1}), \text{ s.t. (8)}. \quad (11)$$

Under some mild conditions, weak duality and strong duality exist between the primal and dual problems we have derived. By strong duality, the two problems are equivalent. The solution of the dual gives the solution to the primal.

Note that it is critically important to keep two variables $\boldsymbol{w}$ and $\boldsymbol{\rho}$ to arrive at the dual (11). One may obtain a different formulation otherwise, and no column generation based optimization can be obtained.

In our case, $A$ is semidefinite but not strictly positive definite, and its inverse does not exist. We can replace its inverse $A^{-1}$ with the Moore-Penrose pseudo-inverse $A^\dagger$. In our experiments, we have regularized $A$ by $A = A + \delta \mathbf{I}$, where $\mathbf{I}$ is the identity matrix and $\delta$ is a small constant.

It is now clear that the dual problem (11) is a regularized hard-margin LPBoost. The second term in the cost function regularizes the dual variable $\boldsymbol{u}$. For example, when $A$ is the identity matrix, this regularization term encourages $\boldsymbol{u}$ to approach $\mathbf{1}$. Also note that here $\boldsymbol{u}$ can take any value and it is not a distribution any more. In contrast, in AdaBoost and LPBoost, $\boldsymbol{u}$ is a distribution: $\boldsymbol{u} \succcurlyeq 0$ and $\mathbf{1}^\top \boldsymbol{u} = 1$.

The Bayesian interpretation of norm-based regularization is as follows. $\ell_2$-norm assumes a Gaussian prior probability over the parameter, and $\ell_1$-norm assumes a Laplacian prior probability. If we view the regularization term as the log of the probability for the parameter $\boldsymbol{x}$, we have

$$- \log p(\boldsymbol{x}) = \begin{cases} \boldsymbol{x}^\top A^{-1}\boldsymbol{x}, & \text{if } p(\boldsymbol{x}) = \mathcal{G}(\mathbf{0}, A), \\ \|\boldsymbol{x}\|_1, & \text{if } p(\boldsymbol{x}) = \prod_i \exp|x_i|, \end{cases} \quad (12)$$

where $\mathcal{G}(\mathbf{0}, A)$ is a Gaussian distribution with zero mean and covariance $A$. In practice, a zero-mean and unit-variance Gaussian prior is usually assumed for kernel ridge regression, while a Laplacian prior over coefficients is typically used in sparse coding and compressed sensing.

In our case, when the number of training data is large ($M \gg 1$), $A$ can be approximated by the identity matrix. The regularization term is simply the variance of the weights associated with each datum. Intuitively, one can design $A$ which contains useful prior information for some particular purpose.

### A. Column Generation Based Optimization

With the above analysis, a column generalization based technique is ready to solve the problem (5).

Instead of solving (5) directly, one calculates the most violated constraint in (11) iteratively for the current solution and adds this constraint to the optimization problem. In theory, any column that violates dual feasibility can be added. To speed up the convergence, we add the most violated constraint by solving the following problem:

$$h'(\cdot) = \underset{h(\cdot)}{\operatorname{argmax}} \; \sum_{i=1}^M u_i y_i h(\boldsymbol{x}_i). \quad (13)$$

This is actually the same as the one that standard AdaBoost and LPBoost use for producing the best weak classifier. That is

---

**Algorithm 1**: Column generation based MDBoost.

**Input**: Labeled data $(\boldsymbol{x}_i, y_i)$, $i = 1 \cdots M$; termination
threshold $\varepsilon > 0$; regularization parameter $D$; maximum
number of iterations $N_{\max}$.

**Initialization**: $N = 0$; $\boldsymbol{w} = \boldsymbol{0}$; and $u_i = \frac{1}{M}$, $i = 1 \cdots M$.

**for** iteration $= 1 : N_{\max}$ **do**

   1) Obtain a new base $h'(\cdot)$ by solving (13);

   2) Check for optimal solution:
     **if** iteration $> 1$ and $\sum_{i=1}^{M} u_i y_i h'(\boldsymbol{x}_i) < r + \varepsilon$,
     **then** break; and the problem is solved;

   3) Add $h'(\cdot)$ to the restricted master problem, which
     corresponds to a new constraint in the dual;

   4) Solve the dual problem (11) and update $r$ and $u_i$
     $(i = 1 \cdots M)$.

   5) Count weak classifiers $N = N + 1$.

**end**

**Output**:

   1) Compute the primal variable $\boldsymbol{w}$ from the optimality
     conditions and the last solved dual problem (primal-dual
     interior point methods output $\boldsymbol{w}$ as well);

   2) The final strong classifier is $F(\boldsymbol{x}) = \sum_{j=1}^{N} w_j h_j(\boldsymbol{x})$.

---

to say, to find the weak classifier that has minimum weighted training error. We summarize our MDBoost in Algorithm 1.

The convergence of Algorithm 1 is guaranteed by general column generation or cutting-plane algorithms, which is easy to establish. When a new $h'(\cdot)$ that violates dual feasibility is added, the new optimal value of the dual problem (maximization) would decrease. Accordingly, the optimal value of its primal problem decreases too because they have the same optimal value due to zero duality gap. Moreover the primal cost function is convex, therefore in the end it converges to the global minimum. Note that on the last step of the proposed MDBoost algorithm is that we can get the value of $\boldsymbol{w}$ easily. Primal-dual interior-point (PD-IP) methods work on the primal and dual problems simultaneously and therefore both primal and dual variables are available after convergence. MDBoost is *totally-corrective* in the sense that the coefficients of all weak classifiers are updated at each iteration.

## IV. EXPERIMENTS

In this section, we run experiments to show the effectiveness of the proposed MDBoost algorithm. In order to control the complexity of the classifier, we use the stumps as weak classifiers.

We first show some results on a synthetic dataset. 800 2D points are generated as shown in Fig. 1 (top). 60% is used for training and the remaining for testing. We then run AdaBoost (1000 iterations) and MDBoost ($N_{\max} = 1000$) on this dataset. The cumulative margin distribution is plotted in Fig. 1 (middle). We have set the parameter $D$ as the sum of weak classifiers' coefficients of AdaBoost. In this experiment, MDBoost's average margin is very similar to AdaBoost's average margin (both are 0.9). However, as observed, the variance of MDBoost is smaller than that of AdaBoost (0.027 *vs.* 0.039). MDBoost also performs slightly better than AdaBoost (3.8% *vs.* 5.0% in test error). Note that, in terms of the minimum margin, AdaBoost is better than MDBoost. This confirms that the minimum margin is not a direct measure of test error. In Fig. 1 (bottom) we show the normalized value of $\boldsymbol{w}$ of the final selected weak classifiers for both algorithms. It can be seen that both algorithms select very similar decision stumps. However, the weights could be different. In Fig. 1 (bottom), the $x$-axis is the index of all the 2500 candidate weak classifiers. If a weak classifier is not selected, then its corresponding weight ($y$-axis) is zero.

Secondly, in order to provide a clearer insight into the feature selection of AdaBoost and MDBoost, we implement AdaBoost and MDBoost on the UCI dataset **spam** whose features have explicit meanings. The task is to separate spam emails based on word frequencies. The training iterations are both constrained as below 60, which is close to the dimension of feature space. We repeat the experiments for 20 times and record the frequency of each feature (word) being selected by the boosting algorithms. The average frequencies over 20 rounds are shown as a histogram in Fig. 2. Note that there is a cross validation (candidates for the parameter $D$ are $\{2, 5, 8, 10, 12, 15, 20, 30, 40, 50, 70, 90, 100, 120\}$) for MDBoost. For AdaBoost, since the best test error is achieved before 60 iterations and no over-fitting is observed during training, the classifier obtained at iteration 60 is considered as optimal. As is illustrated in Fig. 2, both algorithms select important features such as "free" (feature #16 on the plot), "hp" (25), and "$" (53) with high frequencies. However, for the other features, two algorithms select them with diverse inclinations. MDBoost tends to select the features like "address" (15), "order" (9) and "000" (23) which are intuitively helpful for the classification. On the contrary, the favorite ones of AdaBoost, such as "report" (14), "email" (18) and "conference" (48) are more irrelevant for spam email detection. The fact that MDBoost has smaller average test error ($11.3\% \pm 1.20\%$) than AdaBoost ($12.2\% \pm 1.55\%$) supports our analysis.

In the third experiment, we run MDBoost on real datasets and we focus on comparing test error. We have compared four boosting algorithms, which are standard AdaBoost, soft-margin LPBoost [10], AdaBoost-CG[4] and MDBoost, respectively.

The cross validation values for the parameter $D$ for MD-Boost and AdaBoost-CG are $\{2, 5, 8, 10, 12, 15, 20, 30, 40, 50, 70, 90, 100, 120\}$. The trade-off parameter $C$ for LPBoost [10] are $\{0.001, 0.002, 0.005, 0.007, 0.01, 0.02, 0.03, 0.05, 0.07, 0.1, 0.15, 0.2, 0.3, 0.4, 0.5\}$. The experiments are run on the 13 UCI benchmark datasets from [19][5]. Generally, we randomly split the dataset into 3 subsets. 60% of the examples are used for training; 20% are used for cross validation and the other 20% are used for test. For those large datasets (**ringnorm**, **twonorm** and **waveform**), due to the large size, we randomly select 10% for training, 30% for cross validation and 60% for test. For these 3 large datasets, we repeat the experiments for 10 times due to the datasets' large sizes. For the other 10 datasets, experiments are run for 50 times.

The convergence threshold $\varepsilon$ for LPBoost, AdaBoost-CG

---

[4]AdaBoost-CG is a totally corrective version of AdaBoost. It solves $\min_{\boldsymbol{w}} \log(\sum_{i=1}^{M} \exp(-y_i H_i \cdot \boldsymbol{w}))$, s.t. $\boldsymbol{w} \succcurlyeq \boldsymbol{0}, \boldsymbol{1}^\top \boldsymbol{w} = D$ using column generation. See [15] for details.

[5]http://ida.first.fraunhofer.de/projects/bench/

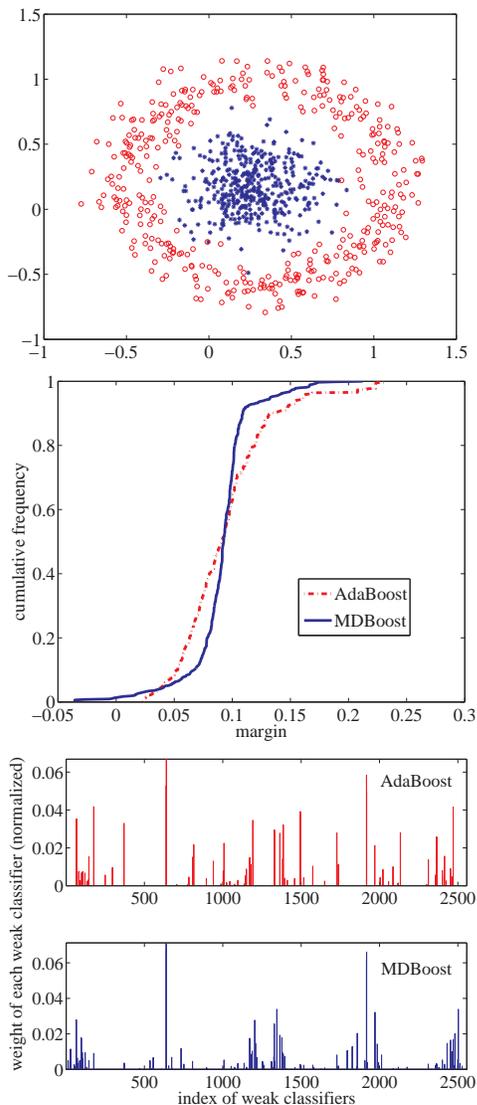

Fig. 1. Toy data: (top) the data; (middle) the cumulative frequency of margin distributions; (bottom) the normalized value of $\boldsymbol{w}$ of the final learned weak classifiers.

and MDBoost are all set to $10^{-5}$. Both test and training results for the four compared algorithms are reported in Table I for a maximum number of iterations of 100, 500 and 1000. In some cases, the three totally-corrective boosting algorithms (LPBoost, AdaBoost-CG, MDBoost) converges earlier than 100 iterations. We simply copy the converged results to iteration 500 and 1000 as reported in Table I.

As can be seen, in terms of training error, soft-margin LPBoost demonstrates its fastest convergence in the training procedure. It finishes the column generation iteration procedure within 100 rounds for 12 datasets but only defeats other algorithms on training error for one dataset (**banana**). On the other hand, the standard AdaBoost, because of its coordinate descent optimization strategy, converges slowest on all the datasets but ranks the best on training error for 10 datasets. MDBoost ties with the LPBoost on training error comparison while AdaBoost-CG outperforms on 2 data sets.

In terms of test error, the proposed MDBoost outperforms on most datasets (9 among 13) and could be considered the best algorithm with respect to the generalization error. The quantitative analysis for the superiority of MDBoost will be discussed later. AdaBoost-CG has the best performance on 3 datasets. The standard AdaBoost wins the remaining one (**thyroid**). It is surprising to find that the LPBoost performs *slightly* worse than the other algorithms on all datasets. It has been observed that different LP solvers may result in slightly different performances on test data for LPBoost [9]. On some datasets, there is a significant difference between PD-IP and simplex based solvers in terms of iterations and the final selected weak classifiers. Here we have used Mosek [20], which implements PD-IP methods. Experiments with simplex LP solvers are needed to verify the LPBoost results. We leave this as future work. In summary, the proposed MDBoost algorithm shows competitive classification performance over AdaBoost, LPBoost and AdaBoost-CG. This validates the usefulness of optimizing margin distributions.

In terms of computational complexity, at each iteration, MDBoost needs to solve a convex QP. The complexity of solving a QP is slightly worse than solving an LP, and it is still very efficient. Moreover, those techniques developed for solving large-scale support vector machines may be applicable here. AdaBoost-CG needs to solve a general convex problem at each iteration, which is much slower than solving a QP or LP [15].

In order to verify the superior classification performance of the proposed MDBoost quantitatively, we implement three statistical comparisons, namely Wilcoxon signed-rank test, Friedman test and Bonferroni-Dunn test, respectively [21], on the experimental performances of the 4 compared boosting algorithms.

The Wilcoxon signed-ranks test (WSRT) [21] is a non-parametric alternative of the paired t-test, which ranks the difference in performance of two classifiers for each dataset. Here the WSRT is used for comparing MDBoost with the other 3 boosting approaches in terms of classification performance. The null-hypothesis declares that the concerning classifier is *no better than* the other algorithms on performance. Consequently, it is a one-tail test with a conventional confidence level (of 95% in this work). Further details of WSRT is illustrated in Table II. As shown, the null-hypothesis is rejected in the tests of MDBoost *vs.* AdaBoost, MDBoost *vs.* AdaBoost-CG and MDBoost *vs.* LPBoost. In other words, MDBoost is considered superior to the other 3 boosting algorithms with respect to generalization error. AdaBoost-CG is the second best since it defeats AdaBoost and LPBoost in the test. AdaBoost and LPBoost could not be considered as better than any other algorithms. To conclude, WSRT indicates that, *with a confidence level of 95%, MDBoost is the best classifier on the 13 datasets used.*

Friedman test (FT) is a non-parametric equivalent of the repeated-measures ANOVA (Analysis of Variance) [21]. FT can measure the difference between more than two sets of classification results. If the null-hypothesis, which assumes that all the performances are similar to each other, is rejected, a *post-hoc* test is processed to compared the algorithms pairwisely. The Bonferroni-Dunn test (BDT) [21] is then adopted as our post-hoc manner to verify whether a classifier over-performs the others under the circumstance of multiple-comparison. Not surprisingly, FT rejects its null-hypothesis, which means the performances of the 4 boosting approaches are different essentially. The confidence level for this test is also set to 95%.[6] We then run BDT. The results of BDT are reported in Table III. According to BDT, different from WRST, only MDBoost statistically significantly outperforms both AdaBoost and LPBoost in test error. Also AdaBoost-CG is not significantly better than AdaBoost with this FT comparison. The conclusion that we can draw here is: *with a confidence level of 95%, (1) MDBoost outperforms AdaBoost and LPBoost; (2) MDBoost and AdaBoost-CG are not significantly different statistically.*

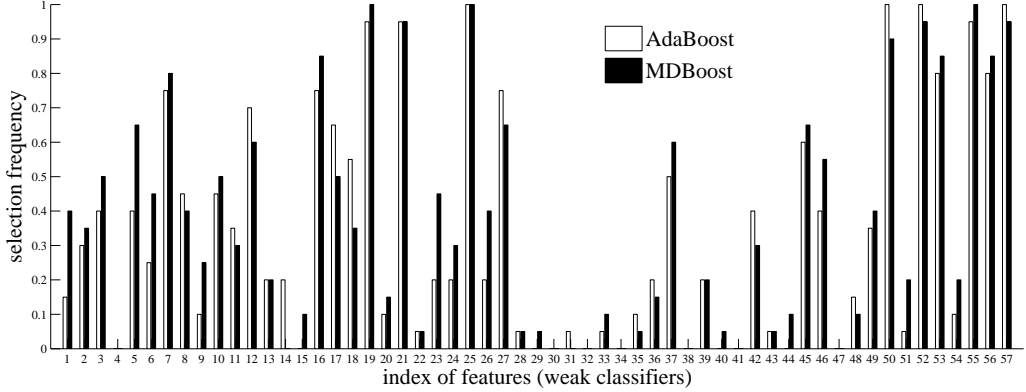

Fig. 2.    The frequencies of different features being selected on the **spam** dataset. Both algorithms select important features such as "free", "hp", and "$" with high frequencies.

To take a close look at the convergence behavior of MD-Boost, we plot the training and test error of AdaBoost and MDBoost for 3 datasets in Fig. 3. Typically, MDBoost converges faster than AdaBoost due to its totally-corrective update rule. In terms of test error, MDBoost is better in most cases as we have reported. On the **breast-cancer** dataset, clearly AdaBoost over-fits the training data. Also both MDBoost and AdaBoost plateau before reaching zero on **banana** because at some point in the algorithms, decision stumps are not able to provide better error rates anymore.

## V. Conclusion

In this paper, we have proposed a new boosting method that optimizes the margin distribution by maximizing the average margin and at the same time minimizing the margin variance. Inspired by LPBoost [10], a column generation based optimization algorithm is proposed to facilitate this idea.

The proposed MDBoost inherits LPBoost's advantages such as well-defined convergence criteria, fast convergence rates and less number of weak learners in the final strong classifier.

Our experiments on various datasets show that MDBoost outperforms AdaBoost and LPBoost; and is at least equivalent to (if not better than) AdaBoost-CG in terms of classification accuracy.

A future research direction is how to integrate useful prior information into the matrix $A$. We believe that improved performance may be obtained by carefully designing $A$. For example, one can take asymmetric data distribution into consideration by giving a weight to each margin $\rho_i$, $i = 1 \cdots M$. We also want to explore the robustness of MDBoost. Since MDBoost considers the whole margin distribution, it is supposed to be more robust to outliers. More experiments are required to test this issue.

## Appendix

The proof of Theorem 2.1 can be found in [15]. For self-completeness, we include the main sketch of the proof here.

A lemma is needed first.

**Lemma 5.1.** *The margin of AdaBoost follows the Gaussian distribution. In general, the larger the number of weak classifiers, the more closely does the margin follow the form of Gaussian under the assumption that selected weak classifiers are uncorrelated.*

We omit the complete proof for this lemma here. The main tool is the central limit theorem. A condition for applying the central limit theorem is that the $N$ variables must be independent. In our case, loosely speaking, AdaBoost selects *independent* weak classifiers such that each weak classifier makes different errors on the training dataset. This may not always hold in practice, but can be a reasonable approximation. So approximately we can view the selected weak classifiers are uncorrelated.

We know that the cost function that AdaBoost minimizes is

$$f(\boldsymbol{w}) = \log\left(\sum_{i=1}^{M} \exp{-\rho_i}\right) \qquad (14)$$

where $\rho_i$ is unnormalized margin for datum $\boldsymbol{x}_i$. As shown in the above lemma, $\rho_i$ follows a Gaussian

$$\mathcal{G}(\rho; \mu, \sigma) = \frac{1}{\sqrt{2\pi}\sigma} \exp{-\frac{(\rho - \mu)^2}{2\sigma^2}},$$

with mean $\mu$ and variance $\sigma^2$.

---

[6] In this case, the critical value equals to 2.291 with the number of classifiers being 4.

TABLE I
TEST AND TRAINING ERRORS (IN PERCENTAGE %) OF ADABOOST (AB), ADABOOST-CG (AB-CG), LPBOOST (LP) AND MDBOOST (MD). FOR
DATASETS **twonorm**, **ringnorm** AND **waveform**, WE RUN THE EXPERIMENTS FOR 10 TIMES DUE TO THE DATASETS' LARGE SIZES. FOR ALL THE OTHERS,
EXPERIMENTS ARE RUN FOR 50 TIMES. THE MEAN AND STANDARD DEVIATION ARE REPORTED. WE HAVE USED DECISION STUMPS AS WEAK
CLASSIFIERS. IN MOST CASES, MDBOOST OUTPERFORMS ADABOOST AND LPBOOST.

| | | test error 100 | test error 500 | test error 1000 | train error 100 | train error 500 | train error 1000 |
|---|---|---|---|---|---|---|---|
| **banana** | AB | 28.5 ± 1.4 | 28.1 ± 1.0 | 28.0 ± 1.1 | 25.9 ± 1.1 | 24.7 ± 0.7 | 24.1 ± 0.7 |
| | AB-CG | 28.0 ± 1.0 | 28.0 ± 1.0 | 28.0 ± 1.0 | 24.9 ± 1.0 | 24.9 ± 1.1 | 24.9 ± 1.1 |
| | LP | 37.9 ± 5.6 | 33.0 ± 1.9 | 32.2 ± 2.1 | 34.8 ± 5.1 | **14.2 ± 1.7** | **7.9 ± 7.8** |
| | MD | **27.7 ± 0.6** | **27.7 ± 0.7** | **27.7 ± 0.7** | **22.6 ± 2.4** | 21.9 ± 3.3 | 21.9 ± 3.3 |
| **b-cancer** | AB | 30.9 ± 5.4 | 31.9 ± 5.5 | 32.3 ± 5.6 | 19.9 ± 2.1 | 20.1 ± 1.7 | 20.3 ± 1.6 |
| | AB-CG | 29.4 ± 5.7 | 29.4 ± 5.7 | 29.4 ± 5.7 | 20.4 ± 2.0 | 20.4 ± 2.0 | 20.4 ± 2.0 |
| | LP | 34.0 ± 7.2 | 34.0 ± 7.2 | 34.0 ± 7.2 | 24.7 ± 4.0 | 24.7 ± 4.0 | 24.7 ± 4.0 |
| | MD | **28.5 ± 4.4** | **28.5 ± 4.4** | **28.5 ± 4.4** | **19.8 ± 2.3** | **19.8 ± 2.3** | **19.8 ± 2.3** |
| **diabetes** | AB | 24.4 ± 3.4 | 26.1 ± 3.8 | 26.8 ± 3.6 | 15.9 ± 1.2 | **9.5 ± 1.2** | **5.0 ± 1.2** |
| | AB-CG | 24.5 ± 3.7 | 24.5 ± 3.7 | 24.5 ± 3.7 | 15.5 ± 5.4 | 15.5 ± 5.6 | 15.5 ± 5.6 |
| | LP | 26.7 ± 4.4 | 26.4 ± 3.7 | 26.4 ± 3.7 | **11.6 ± 4.0** | 11.1 ± 4.5 | 11.1 ± 4.5 |
| | MD | **23.8 ± 4.0** | **23.7 ± 3.9** | **23.7 ± 3.9** | 17.0 ± 4.0 | 17.0 ± 4.2 | 17.0 ± 4.2 |
| **f-solar** | AB | 34.2 ± 3.5 | 35.2 ± 3.7 | 35.3 ± 3.7 | 33.1 ± 1.6 | 33.2 ± 1.7 | 33.2 ± 1.7 |
| | AB-CG | 34.0 ± 3.4 | 34.0 ± 3.4 | 34.0 ± 3.4 | **32.9 ± 1.5** | **32.9 ± 1.5** | **32.9 ± 1.5** |
| | LP | 34.1 ± 3.7 | 34.1 ± 3.7 | 34.1 ± 3.7 | 33.3 ± 1.6 | 33.3 ± 1.6 | 33.3 ± 1.6 |
| | MD | **34.0 ± 3.5** | **34.0 ± 3.5** | **34.0 ± 3.5** | 32.9 ± 1.6 | 32.9 ± 1.6 | 32.9 ± 1.6 |
| **german** | AB | 25.3 ± 2.9 | 26.6 ± 2.8 | 27.6 ± 2.8 | 18.4 ± 1.1 | **15.7 ± 1.2** | **13.9 ± 1.2** |
| | AB-CG | **25.6 ± 3.0** | **25.5 ± 3.0** | **25.5 ± 3.0** | 18.4 ± 2.6 | 18.2 ± 3.3 | 18.2 ± 3.3 |
| | LP | 29.9 ± 3.5 | 30.5 ± 3.5 | 30.5 ± 3.5 | 21.4 ± 8.0 | 15.8 ± 12.8 | 15.8 ± 12.8 |
| | MD | 25.7 ± 2.9 | 25.6 ± 2.8 | 25.6 ± 2.8 | **17.5 ± 3.0** | 17.4 ± 3.2 | 17.4 ± 3.2 |
| **heart** | AB | 19.4 ± 5.0 | 21.4 ± 5.8 | 22.0 ± 5.8 | **3.1 ± 1.3** | **0 ± 0** | **0 ± 0** |
| | AB-CG | 17.1 ± 4.7 | 17.1 ± 4.7 | 17.1 ± 4.7 | 10.7 ± 3.0 | 10.7 ± 3.0 | 10.7 ± 3.0 |
| | LP | 18.5 ± 5.7 | 18.5 ± 5.7 | 18.5 ± 5.7 | 9.4 ± 5.9 | 9.4 ± 5.9 | 9.4 ± 5.9 |
| | MD | **16.1 ± 4.2** | **16.1 ± 4.2** | **16.1 ± 4.2** | 10.6 ± 3.1 | 10.6 ± 3.1 | 10.6 ± 3.1 |
| **image** | AB | 4.2 ± 0.9 | 3.0 ± 0.8 | 2.9 ± 0.9 | 2.4 ± 0.4 | **0 ± 0** | **0 ± 0** |
| | AB-CG | **3.1 ± 0.9** | **3.1 ± 0.9** | **3.1 ± 0.9** | **0.2 ± 0.4** | 0.2 ± 0.4 | 0.2 ± 0.4 |
| | LP | 3.7 ± 1.2 | 3.2 ± 0.9 | 3.2 ± 0.9 | 0.8 ± 0.8 | 0.8 ± 0.8 | 0.8 ± 0.8 |
| | MD | 3.6 ± 0.9 | 3.3 ± 1.0 | 3.3 ± 1.0 | 1.8 ± 0.5 | 1.7 ± 0.6 | 1.7 ± 0.6 |
| **ringnorm** | AB | 7.5 ± 0.6 | 5.3 ± 0.4 | 5.2 ± 0.3 | 1.4 ± 0.5 | **0 ± 0** | **0 ± 0** |
| | AB-CG | 7.3 ± 1.1 | 5.3 ± 0.3 | 5.3 ± 0.3 | **0 ± 0** | **0 ± 0** | **0 ± 0** |
| | LP | 8.3 ± 4.6 | 5.4 ± 0.3 | 5.4 ± 0.3 | 0.6 ± 0.4 | 0.3 ± 0.3 | 0.3 ± 0.3 |
| | MD | **7.2 ± 1.0** | **5.1 ± 0.4** | **5.1 ± 0.4** | 0.9 ± 0.3 | 0.3 ± 0.3 | 0.3 ± 0.3 |
| **splice** | AB | 9.2 ± 1.6 | 1 ± 1.7 | 10.4 ± 1.6 | **5.2 ± 1.5** | **2.9 ± 2.7** | **2.7 ± 2.8** |
| | AB-CG | 8.9 ± 1.3 | 8.9 ± 1.3 | 8.9 ± 1.3 | 6.1 ± 1.5 | 6.1 ± 1.5 | 6.1 ± 1.5 |
| | LP | 9.9 ± 2.0 | 10.2 ± 2.3 | 10.2 ± 2.3 | 8.1 ± 2.1 | 8.0 ± 2.2 | 8.0 ± 2.2 |
| | MD | **8.2 ± 1.0** | **8.2 ± 1.0** | **8.2 ± 1.0** | 6.2 ± 0.6 | 6.2 ± 0.5 | 6.2 ± 0.5 |
| **thyroid** | AB | **6.7 ± 4.3** | **7.3 ± 4.2** | **7.2 ± 4.2** | **0 ± 0** | **0 ± 0** | **0 ± 0** |
| | AB-CG | 7.8 ± 4.5 | 7.8 ± 4.5 | 7.8 ± 4.5 | 1.2 ± 2.0 | 1.2 ± 2.0 | 1.2 ± 2.0 |
| | LP | 7.8 ± 5.1 | 7.8 ± 5.1 | 7.8 ± 5.1 | 1.3 ± 2.0 | 1.3 ± 2.0 | 1.3 ± 2.0 |
| | MD | 7.6 ± 4.9 | 7.6 ± 4.9 | 7.6 ± 4.9 | 1.8 ± 1.5 | 1.8 ± 1.5 | 1.8 ± 1.5 |
| **titanic** | AB | **21.8 ± 1.7** | **21.8 ± 1.7** | **21.8 ± 1.7** | **22.4 ± 0.6** | **22.4 ± 0.6** | **22.4 ± 0.6** |
| | AB-CG | **21.8 ± 1.7** | **21.8 ± 1.7** | **21.8 ± 1.7** | **22.4 ± 0.6** | **22.4 ± 0.6** | **22.4 ± 0.6** |
| | LP | 21.9 ± 1.7 | 21.9 ± 1.7 | 21.9 ± 1.7 | 22.5 ± 1.0 | 22.5 ± 1.0 | 22.5 ± 1.0 |
| | MD | **21.8 ± 1.7** | **21.8 ± 1.7** | **21.8 ± 1.7** | 22.5 ± 0.6 | 22.5 ± 0.6 | 22.5 ± 0.6 |
| **twonorm** | AB | 4.2 ± 0.4 | 4.0 ± 0.4 | 4.0 ± 0.4 | **0 ± 0** | **0 ± 0** | **0 ± 0** |
| | AB-CG | 4.2 ± 0.4 | 4.1 ± 0.4 | 4.1 ± 0.4 | 0 ± 0.1 | 0 ± 0.1 | 0 ± 0.1 |
| | LP | 4.4 ± 0.3 | 4.3 ± 0.3 | 4.3 ± 0.3 | 0.8 ± 0.7 | 0.9 ± 0.7 | 0.9 ± 0.7 |
| | MD | **3.6 ± 0.2** | **3.5 ± 0.2** | **3.5 ± 0.2** | 0.8 ± 0.4 | 0.8 ± 0.3 | 0.8 ± 0.3 |
| **waveform** | AB | 12.5 ± 0.8 | 13.3 ± 0.8 | 13.6 ± 0.8 | **0.6 ± 0.5** | **0 ± 0** | **0 ± 0** |
| | AB-CG | **12.4 ± 0.9** | **12.4 ± 0.9** | **12.4 ± 0.9** | 2.9 ± 2.0 | 2.9 ± 2.0 | 2.9 ± 2.0 |
| | LP | 12.7 ± 0.7 | 12.7 ± 0.6 | 12.7 ± 0.6 | 5.6 ± 2.1 | 5.6 ± 2.2 | 5.6 ± 2.2 |
| | MD | 13.0 ± 1.1 | 12.8 ± 0.9 | 12.8 ± 0.9 | 4.5 ± 1.7 | 4.5 ± 1.7 | 4.5 ± 1.7 |

It is well known that the Monte Carlo integration method can be used to compute a continuous integral

$$\int g(x) f(x) dx \simeq \frac{1}{K} \sum_{k=1}^{K} f(x_k), \qquad (15)$$

where $g(x)$ is a probability distribution such that $\int g(x) dx = 1$ and $f(x)$ is an arbitrary function. $x_k$, $(k = 1 \cdots K)$, are randomly sampled from the distribution $g(x)$. The more samples are used, the more accurate the approximation is.

Equation (14) can be viewed as a discrete Monte Carlo approximation of the following integral (here the constant term $\log M$ is discarded, which is irrelevant to the analysis):

$$f'(\boldsymbol{w})$$
$$= \log \int_{\rho_1}^{\rho_2} \mathcal{G}(\rho; \mu, \sigma) \exp(-\rho) \, d\rho$$
$$= \log \int_{\rho_1}^{\rho_2} \frac{1}{\sqrt{2\pi}\sigma} \exp\left(-\frac{(\rho - \mu)^2}{2\sigma^2} - \rho\right) d\rho$$

TABLE II

RESULT OF WILCOXON SIGNED-RANKS TEST (WSRT). THE TEST IS PERFORMED PAIRWISELY AMONG ADABOOST, ADABOOST-CG, LPBOOST AND MDBOOST. THE BLOCK WHERE "BETTER" TAKES PLACE INDICATES THAT THE ALGORITHM CORRESPONDING TO ITS ROW IS BETTER THAN THE ALGORITHM CORRESPONDING TO ITS COLUMN WHILE "NOT BETTER" SUGGESTS THE CONTRARY. NUMBERS IN THE PARENTHESES ARE THE NUMERICAL RESULTS OF WSRT $z$ (THE LARGER THE BETTER). NOTE THAT THE STATEMENT "BETTER" ONLY TAKES PLACE WHERE $z$ IS LARGER THAN THE CRITICAL VALUE. THE CRITICAL VALUE DEPENDS ON THE NUMBER OF DATASETS THAT HAVE DIFFERENCE CLASSIFICATION PERFORMANCES. HENCE IT IS NOT A FIXED NUMBER. IN OUR CASE, THERE ARE TWO CRITICAL VALUES: $v_1 = 61$ AND $v_2 = 70$ [21]. THOSE MARKED WITH * SHOULD BE COMPARED AGAINST $v_1$ AND OTHERS AGAINST $v_2$.

|  | AdaBoost | AdaBoost-CG | LPBoost | MDBoost |
|---|---|---|---|---|
| **AdaBoost** | – | Not Better (12*) | Not Better (43) | Not Better (7) |
| **AdaBoost-CG** | **Better** (66*) | – | **Better** (78*) | Not Better (18) |
| **LPBoost** | Not Better (48) | Not Better (0*) | – | Not Better (5) |
| **MDBoost** | **Better** (84) | **Better** (73) | **Better** (86) | – |

TABLE III

RESULTS OF BONFERRONI-DUNN TEST (BDT). EACH ALGORITHMS IS COMPARED WITH OTHER 3 BOOSTING MANNERS AT THE SAME TIME. THE BLOCK WHERE "BETTER" TAKES PLACE INDICATES THAT THE ALGORITHM CORRESPONDING TO ITS LINE IS BETTER THAN THE ALGORITHM CORRESPONDING TO ITS COLUMN WHILE "NOT BETTER" SUGGESTS THE CONTRARY. NUMBERS IN THE PARENTHESES ARE THE NUMERICAL RESULTS OF BDT $z$ (THE LARGER THE BETTER). NOTE THAT THE STATEMENT "BETTER" ONLY TAKES PLACE WHERE $z$ IS LARGER THAN THE CRITICAL VALUE, WHICH IS 2.291.

|  | AdaBoost | AdaBoost-CG | LPBoost | MDBoost |
|---|---|---|---|---|
| **AdaBoost** | – | Not Better ($-1.75$) | Not Better (0.76) | Not Better ($-2.66$) |
| **AdaBoost-CG** | Not Better (1.75) | – | **Better** (2.51) | Not Better ($-0.91$) |
| **LPBoost** | Not Better ($-0.76$) | Not Better ($-2.51$) | – | Not Better ($-3.42$) |
| **MDBoost** | **Better** (2.66) | Not Better (0.91) | **Better** (3.42) | – |

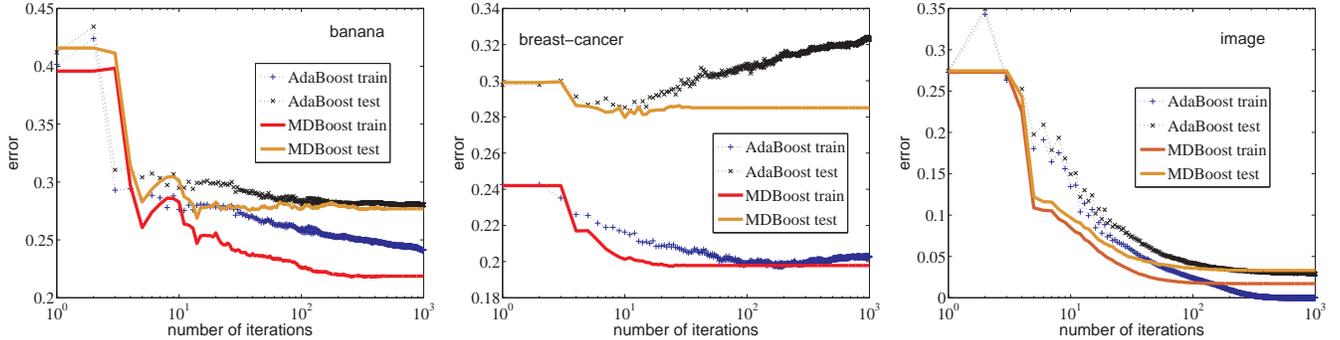

Fig. 3.   Training error and test error of AdaBoost and MDBoost for **banana**, **breast-cancer**, **image** datasets. These convergence plots correspond to the results in Table I.

$$= \log\left[\frac{1}{2}\exp\left(-\mu + \frac{\sigma^2}{2}\right)\text{Erf}\left(\frac{\rho - \mu}{\sqrt{2}\sigma} + \frac{\sigma}{\sqrt{2}}\right)\Big|_{\rho_1}^{\rho_2}\right]$$
$$= -\log 2 - \mu + \frac{\sigma^2}{2} + \log\left[\text{Erf}\left(\frac{\rho - \mu}{\sqrt{2}\sigma} + \frac{\sigma}{\sqrt{2}}\right)\Big|_{\rho_1}^{\rho_2}\right],$$
$$(16)$$

where $\text{Erf}(x) = \frac{2}{\sqrt{\pi}}\int_0^x \exp{-s^2}ds$ is the Gauss error function. The integral range is $[\rho_1, \rho_2]$. We do not know explicitly about the integration range. We can roughly calculate the integral from $-\infty$ to $+\infty$. Then the last term in (16) is $\log 2$ and the result is simple

$$f'(\boldsymbol{w}) = -\mu + \frac{1}{2}\sigma^2. \qquad (17)$$

This approximation is reasonable because Gaussian distributions drop off quickly and Gaussian is not considered a heavy-tailed distribution.

Hence, AdaBoost approximately maximizes the cost function

$$-f'(\boldsymbol{w}) = \mu - \frac{1}{2}\sigma^2. \qquad (18)$$

This cost function has a clear and elegant explanation: The first term $\mu$ is the unnormalized average margin and the second term $\sigma^2$ is the unnormalized margin variance. It is clear that AdaBoost maximizes the unnormalized average margin and also takes minimizing the unnormalized margin variance into account. A better *margin distribution* is then obtained.

### ACKNOWLEDGMENT

The authors thank the anonymous reviewers for their valuable comments, which have significantly improved the quality of this paper.